\documentclass[11pt]{article}

\usepackage[preprint]{acl}

\usepackage{times}
\usepackage{latexsym}

\usepackage[T1]{fontenc}

\usepackage[utf8]{inputenc}

\usepackage{microtype}

\usepackage{sourcecodepro}

\usepackage{graphicx}

\usepackage{booktabs}

\usepackage{multirow}

\usepackage{longtable}

\usepackage[most]{tcolorbox}

\usepackage{upquote}

\usepackage{fvextra}

\usepackage{xspace}

\title{\ourbenchmark: Hierarchical Instruction Following for Large Language Models}

\author{Yuetian Mao \\
  Technical University of Munich \\
  \texttt{yuetian.mao@tum.de} \\\And
  Chunyang Chen \\
  Technical University of Munich \\
  \texttt{chun-yang.chen@tum.de} \\}

\newcommand{\eg}{\hbox{e.g.}\xspace}

\newcommand{\ex}[1]{\textit{``#1''}}
\newcommand{\extype}[1]{\textit{#1}}

\newcommand{\ourbenchmark}{IFHierBench\xspace}
\newcommand{\totalconstraintamount}{35\xspace}
\newcommand{\structureconstraintamount}{10\xspace}
\newcommand{\valueconstraintamount}{25\xspace}
\newcommand{\dataamount}{600\xspace}
\newcommand{\perdepthdataamount}{150\xspace}

\newcommand{\analysismodel}{GPT-5.4\xspace}

\newcommand{\gptmodel}{gpt-5.5-2026-04-23\xspace}
\newcommand{\gptanalysis}{GPT-5.5\xspace}
\newcommand{\claudemodel}{claude-opus-4-6-20251015\xspace}
\newcommand{\claudeanalysis}{Claude Opus 4.6\xspace}
\newcommand{\geminimodel}{gemini-3-flash-preview\xspace}
\newcommand{\geminianalysis}{Gemini-3-Flash\xspace}
\newcommand{\deepseekmodel}{deepseek-r1\xspace}
\newcommand{\deepseekanalysis}{DeepSeek-R1\xspace}
\newcommand{\kimimodel}{kimi-k2.5\xspace}
\newcommand{\kimianalysis}{Kimi-K2.5\xspace}
\newcommand{\qwenmodel}{Qwen3.6-35B-Instruct\xspace}
\newcommand{\qwenanalysis}{Qwen-3.6-35B\xspace}
\newcommand{\gemmamodel}{gemma-4-26b-it\xspace}
\newcommand{\gemmaanalysis}{Gemma-4-26B\xspace}

\begin{document}
\maketitle
\begin{abstract}

Instruction-following ability is critical for deploying large language models in real-world applications, where downstream components depend on the output satisfying specific constraints. Modern deployments increasingly handle the full task in a single LLM call, with one prompt specifying a layered output whose overall artifact, structural sections, and nested fields must each satisfy concrete constraints. Existing instruction-following benchmarks treat the constraint set as a flat list applied uniformly to the response, so they cannot scope a check to a particular section of the output. We introduce \ourbenchmark, a hierarchical instruction-following benchmark of \dataamount prompts stratified across four constraint-tree depths and \totalconstraintamount distinct constraints, each prompt paired with a deterministic checker that verifies satisfaction at every scope. Evaluating seven leading proprietary and open-weight models, we find that even the strongest model only marginally exceeds 50\% prompt-level accuracy and that accuracy degrades sharply as constraint depth grows. Reliably following nested constraints remains a substantial gap for current LLMs, motivating future training methods that consider constraint adherence at finer granularity to achieve better instruction-following ability.
\end{abstract}

\section{Introduction}
Large language models (LLMs) can produce fluent text. Yet semantic correctness alone is not sufficient: they must also follow user-specified instructions. These instructions are often output constraints~\cite{liang2024controllable}, and an empirical study of real-world developer prompts finds that over 80\% of them are objectively verifiable: required fields, format, length, and required or forbidden keywords~\cite{liu2024we}. Such constraints admit deterministic checking and so define a measurable instruction-following task. How users specify these constraints in prompts has changed alongside what LLMs and agent frameworks can support. Early production pipelines were limited by model capacity, short context windows, and immature agent infrastructure, which together forced a full task to be decomposed into many small steps~\cite{khot2022decomposed}.  Each step took a short prompt carrying one or two simple constraints and surrounding application code chained the partial outputs into the larger task. As model capacity and context windows grew and agent frameworks matured~\cite{yang2024swe}, the same task can now be handled in a single LLM call, with the full constraint specification placed inside a long prompt. A single prompt now specifies what the whole response must look like, what each major section inside it must contain, and what each field nested inside those sections must satisfy. Whether current LLMs reliably follow this kind of layered specification is the question we study.

\begin{figure}[t]
  \centering
  \includegraphics[width=\columnwidth]{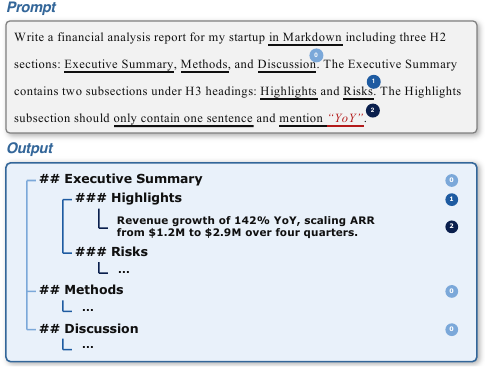}
  \caption{An example of hierarchical instruction following: numbers mark constraint depth, darker boxes denote deeper scopes.}
  \label{fig:markdown-example}
\end{figure}

Consider a prompt asking for a financial analysis report (Figure~\ref{fig:markdown-example}): the whole response must be a Markdown document with three H2 sections, its Executive Summary section must contain two H3 subsections, and the innermost Highlights subsection must be a single sentence mentioning ``YoY''. Real prompts wrap several more such layers around the same task. Existing benchmarks for testing model instruction following treat the constraints in a prompt as a single flat list, with every check applied uniformly to the whole response~\cite{zhou2023instruction, wen2024benchmarking, pyatkin2026generalizing}. Among them, IFEval and IFBench impose independent checks on the whole response, and ComplexBench adds logical relations such as conjunction, chaining, and selection, yet none scope a check to a nested region of the output. On this single-list formulation they are already near saturation, with leading models including ones under 10B parameters scoring above 80\% on IFEval~\cite{yang2025qwen3, guo2025deepseek, zeng2025glm}, while their behavior on prompts with the kind of nested scoping just described remains untested. Treating such a hierarchy as a flat list distorts the failure picture: a flat checker cannot scope a check to a particular section, so it marks the inner constraint ``Highlights mentions YoY'' as satisfied on any response containing the word YoY, even when the outer Executive Summary section that should contain Highlights was never produced.

\begin{table}[t]
\centering
\footnotesize
\renewcommand{\arraystretch}{1.1}
\begin{tabular*}{\columnwidth}{@{\extracolsep{\fill}}llcr@{}}
\toprule
Benchmark & Relation & Verifier & Size \\
\midrule
IFEval & Independent & Code & 541 \\
IFBench & Independent & Code & 300 \\
ComplexBench & Logical & Rule+LLM & 1{,}150 \\
\textbf{\ourbenchmark} & \textbf{Hierarchical} & \textbf{Code} & \textbf{600} \\
\bottomrule
\end{tabular*}
\caption{Comparison of \ourbenchmark with representative instruction-following benchmarks.}
\label{tab:benchmark-comparison}
\end{table}

In this work, we introduce \ourbenchmark, a hierarchical instruction-following benchmark of \dataamount prompts spanning constraint depths 0 through 3, with \perdepthdataamount prompts at each depth. Table~\ref{tab:benchmark-comparison} compares \ourbenchmark with prior instruction-following benchmarks: it is the only one that scopes checks to nested output regions while verifying every check with deterministic code. The benchmark covers \totalconstraintamount distinct constraints. Of these, \structureconstraintamount structure-level constraints over JSON, Markdown, list, and labeled-section scopes are derived from an empirical analysis of output requirements across 1{,}232 real LLM application prompts from GitHub, while the remaining \valueconstraintamount content-level constraints are drawn from IFEval~\cite{zhou2023instruction} and ComplexBench~\cite{wen2024benchmarking}. A data-synthesis pipeline turns each IFEval seed task into a constraint tree through depth-first sampling gated by a sibling-conflict matrix, then renders the tree into a single natural-language prompt by merging compatible sibling constraints into joint sentences. We pair each constraint type with a deterministic Python checker, and the pipeline composes these per-type checkers along the constraint tree to emit a hierarchical checker for every prompt, so that satisfaction at every scope is verified without any model-based judging. Evaluating seven leading models, we find that even the strongest model only marginally exceeds 50\% prompt-level accuracy, and accuracy degrades sharply as constraint depth grows.

Our contributions are listed as follows:
\begin{itemize}
  \item A new hierarchical instruction-following benchmark of \dataamount prompts stratified across four constraint depths, with \totalconstraintamount distinct constraints (\structureconstraintamount structure-level and \valueconstraintamount content-level), where the structure-level set is derived from an empirical analysis of 1{,}232 real LLM application prompts from GitHub. Our code and data are available at \url{https://anonymous.4open.science/r/IFHierBench-0087}.
  \item A data synthesis pipeline that turns real-world tasks into hierarchical constraint trees and renders them as natural-language prompts paired with a deterministic hierarchical checker.
  \item Experiments on seven leading models showing that even the strongest model only marginally exceeds 50\% prompt-level accuracy, and that accuracy degrades sharply as constraint depth grows.
\end{itemize}

\section{Related Work}
\subsection{Controlled Text Generation}
A large body of work has pushed language models to follow user-specified constraints more reliably, through both \emph{decoding-time} biasing and \emph{training-time} fine-tuning. Decoding-time methods either force required tokens or phrases to appear by constraining the beam-search frontier \cite{hokamp2017lexically, post2018fast}, or use auxiliary attribute discriminators and future-aware predictors to bias the next-token distribution at each step \cite{dathathri2019plug,khanov2024args,shin2025eco,yang2021fudge,lu2021neurologic}. Training-time approaches bake constraint adherence into the model itself, by conditioning generation on control codes prepended to the input \cite{keskar2019ctrl,krause2021gedi} or by aligning the model to human preferences through instruction tuning \cite{ouyang2022training}. These methods all operate on \emph{flat, single-scope} constraints that act on the response as a whole. Whether they transfer to the \emph{hierarchical} constraints of real prompts, where a format choice at one scope licenses restrictions on the scopes nested inside it, remains open. \ourbenchmark supplies training and evaluation data in this hierarchical regime.

\subsection{Instruction-Following Benchmarks}
Instruction-following benchmarks evaluate how reliably LLMs satisfy user-specified output constraints. One line enumerates as many distinct constraint types as possible to broadly test instruction-following ability: IFEval~\cite{zhou2023instruction} fixes 25 verifier templates, FollowBench~\cite{jiang2024followbench} stacks five constraint families along a difficulty ladder, IFBench~\cite{pyatkin2026generalizing} adds previously unseen types to test generalization, and CFBench~\cite{zhang2025cfbench} pairs 200+ scenarios with checklist scoring and constraint priorities. A second line studies how multiple constraints interact inside one prompt: ComplexBench~\cite{wen2024benchmarking} attaches explicit logical operators such as conjunction, chain, and selection between constraints. A third line extends instruction-following to multi-turn dialogues~\cite{bai2024mt, kwan2024mt, li2025structflowbench}. None of these benchmarks treats the output of a single turn as a nested structure with constraints scoped to its sub-regions, and most emphasize content-level requirements rather than structure-level correctness. \ourbenchmark fills both gaps by providing prompts whose constraints form a tree over the output's structural geometry, jointly evaluating format-level correctness at outer scopes and content-level constraints at inner scopes.

\section{Structure-level Constraint Taxonomy}

\subsection{Constraint Collection}
\label{sec:data-collection}
Our corpus is built on PromptSet~\cite{pister2024promptset}, a collection of prompts extracted from LLM-based applications (LLMapps) in open-source GitHub projects. These projects span a wide range of use cases and adoption levels, from personal demos to widely deployed systems, so prompt quality varies substantially. To reflect mature practice, we focus on widely used prompt templates (predefined structures that combine static text with dynamic placeholders to create adaptable prompts for LLMapps) in these apps~\cite{schulhoff2024prompt, zhao2025llm}. We then apply the cleaning pipeline of \citet{mao2025prompts} that filters by repository popularity and prompt length, and segment each remaining prompt into different components. We retain only the prompts that contain at least one explicit output constraint. This procedure yields 1{,}232 prompt templates from 2{,}888 prompts in 1{,}525 GitHub repositories.

\subsection{Annotation and Taxonomy}
\label{sec:annotation}
We use GPT-5.4 to annotate each prompt with the structure-level constraint that fixes output format.

We seed the annotator with two output formats, JSON and Markdown (informed by the output format categories in ComplexBench~\cite{wen2024benchmarking}), and allow it to propose a new category whenever no existing one fits. New proposals are reviewed manually and either accepted as a new category or merged into an existing one. Beyond the free-form String case, which covers 50.8\% of the corpus, the four most frequent categories are Labeled Sections (15.0\%), JSON (14.0\%), List (12.7\%), and Markdown (8.9\%), where each percentage is the share of prompts that include that category, and a prompt may declare multiple categories. We adopt these four formats together with String as the host formats of \ourbenchmark: the four structured formats define containers that recursively hold leaf-level value constraints, while String hosts value-level constraints directly without any container scope.

To confirm that hierarchical structure is not merely hypothetical, we measure the nesting depth of the format declaration in each prompt. Among the 606 format-bearing prompts, 411 (67.8\%) carry at least one nested scope, with 255 at depth 1, 131 at depth 2, and 25 at depth 3 or deeper. Beyond depth, 171 of these prompts (28.2\%) declare two or more formats, indicating that nesting in real prompts crosses format boundaries rather than recursing inside a single format. Constraint specifications also take up much of the prompt: across the 1{,}232 prompts they occupy 46.4\% of the text on average. These findings shape \ourbenchmark: the five formats become its host scopes, nesting up to depth three makes each prompt a depth-stratified constraint tree, and the 28.2\% of prompts that mix formats motivate letting a child scope take a different format from its parent.

\section{\ourbenchmark Construction Pipeline}

\begin{figure*}[t]
  \centering
  \includegraphics[width=\textwidth]{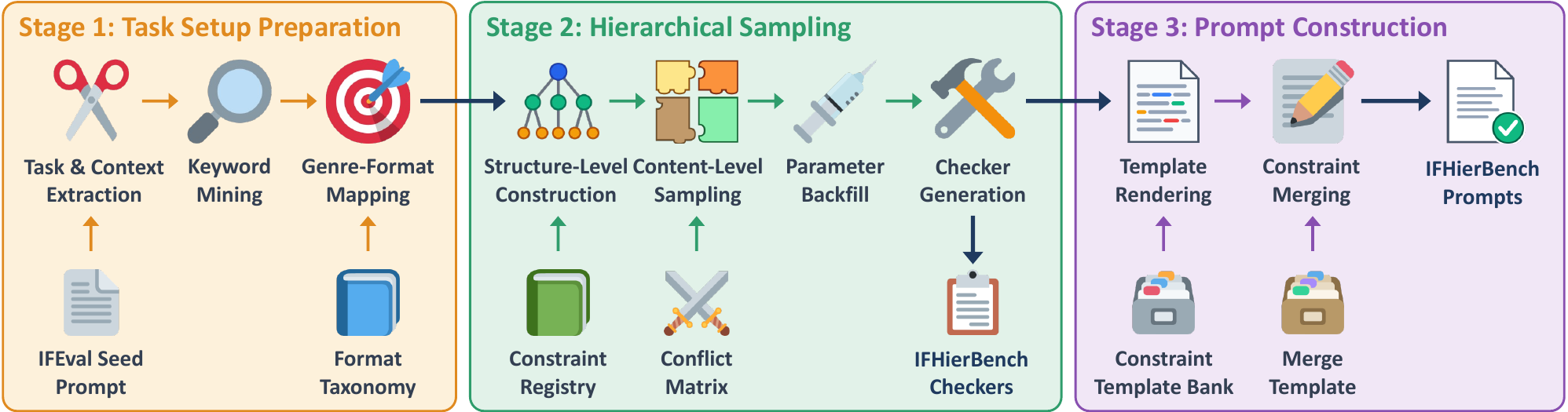}
  \caption{Overview of the \ourbenchmark construction pipeline.}
  \label{fig:pipeline_overview}
\end{figure*}

As shown in Figure~\ref{fig:pipeline_overview}, the \ourbenchmark construction pipeline runs in three stages. \emph{Task setup preparation} parses each seed prompt into a clean task and its context together with a set of task-related keywords, and decides which structure-level constraints are admissible for the task. \emph{Hierarchical sampling} first grows the constraint-tree skeleton from root structure-level constraints, then samples content-level templates at each leaf, and finally resolves parameters across all templates to finalize the tree. An automatic checker is derived from the finalized tree to verify model responses.
\emph{Prompt construction} renders the constraint tree into a single natural-language instruction, resolves cross-scope references, and merges constraints into more compact and readable sentences.

\subsection{Task Setup Preparation}
\label{sec:task-setup}

This stage turns each seed prompt into a sampling-ready task: a clean task statement free of output constraints, the context the task operates on, a task-specific vocabulary pool, and the subset of root formats that are admissible for the task's genre.

\textbf{Task and Context Extraction.} The first step parses each IFEval seed prompt into a task description the model must complete, together with the material it operates on. We use \analysismodel to return a clean task description with all output-related constraints stripped, together with the verbatim context the model operates on. The clean task lets downstream sampling attach new constraints to a goal that no longer carries any of its own, and the verbatim context keeps the rendered prompt self-sufficient under subsequent processing. Our running example (Figure~\ref{fig:setup_example}, panel 1) illustrates this step: the seed \ex{Write a news brief of no more than 80 words based on the following article.} is parsed into the clean task \ex{Write a news brief based on the following article.} with its output constraints stripped, together with the article as the context.

\textbf{Keyword Mining.} Since some constraints restrict the lexical choice of the output (\eg, \extype{requiring the response to include specific keywords}, \extype{forbidding specific keywoprd}, \extype{naming a JSON key}), we use \analysismodel to mine a set of vocabulary pools tailored to each task. These pools feed the sampler in the next stage. Panel 2 of Figure~\ref{fig:setup_example} shows the pool emitted for the running example.

\textbf{Genre-Format Mapping.} Different tasks target different output genres (\eg, \extype{news brief}, \extype{poem}, \extype{email}, \extype{analysis report}), and each genre carries implicit restrictions on the admissible format: a news brief naturally accepts a JSON object, labeled sections, or markdown headings (Figure~\ref{fig:setup_example}, panel 3), whereas an email rejects bulleted lists or JSON wrappers. For each task we ask \analysismodel to extract its genre and then to decide which of our predefined root formats are admissible given that genre, and the sampler draws a root format only from this admissible subset. Because task extraction, keyword mining, and genre-format mapping are each carried out by \analysismodel, we randomly sample 10\% of each step's outputs and manually evaluate them, and every step achieves over 90\% accuracy.

\begin{figure}[t]
\centering
\refstepcounter{figure}\label{fig:setup_example}%
\begin{tcolorbox}[
  colback=orange!12,
  colframe=orange!55!black,
  title={\textbf{Figure~\thefigure: running example for~3.1}},
  fonttitle=\footnotesize,
  boxsep=1pt,left=4pt,right=4pt,top=2pt,bottom=2pt,
  fontupper=\footnotesize
]
{\itshape Seed prompt}\\
``Write a news brief of no more than 80 words based on the following article. [article body omitted]''

\smallskip
{\itshape (1) Task and context}\\
\textbf{task:} ``Write a news brief based on the following article.''\\
\textbf{context:} [article body omitted]

\smallskip
{\itshape (2) Mined vocabulary pool}\\
\textbf{related keywords:} report, policy, briefing, \ldots

\smallskip
{\itshape (3) Genre and admissible root formats}\\
\textbf{genre:} news brief\\
\textbf{admits:} JSON, Markdown, Labeled Sections, String
\end{tcolorbox}
\end{figure}

\subsection{Hierarchical Sampling}
\label{sec:sampling}
Given a prepared task, this stage produces a tree of constraints rooted at the chosen output format, together with the deterministic checker that will later score model responses against it. 

\textbf{Structure-Level Construction.} The sampler walks the tree depth-first. At each scope it first chooses the shape of the scope (\extype{string}, \extype{JSON object}, \extype{list}, \extype{markdown document}, or \extype{labeled section}) and the structure-level constraints that the scope as a whole must satisfy, such as a required JSON key set, a fixed list length, or a mandatory markdown heading. Container scopes recursively spawn child scopes whose kind is resolved through a weighted choice that respects the parent kind and the remaining depth budget, and at the maximum depth children are forced to be atomic strings to avoid empty containers. The running example (top panel of Figure~\ref{fig:tree_example}) illustrates this: a JSON root carries required keys ``headline'' and ``bullets'', ``bullets'' expands into a length-3 list scope, and at the depth boundary each list element is forced to an atomic string. Each scope receives a unique alias when it is opened, so that constraints declared inside it can refer back to it unambiguously from the surface prompt.

\textbf{Content-Level Sampling.} Inside each scope, the sampler draws content-level constraints that further restrict the text produced at that scope, such as \extype{required keywords}, \extype{forbidden words}, \extype{casing patterns}, or \extype{length counts}. A candidate is admissible only if it does not conflict with siblings already drawn for the same scope, since two individually feasible constraints can be jointly unsatisfiable. Inspired by IFBench~\cite{pyatkin2026generalizing}, we maintain a \emph{conflict matrix} that records pairwise incompatibilities between content constraints, and the sampler consults it on every draw. The matrix encodes physical mutex groups, in which at most one member may appear in any scope (for example, a scope cannot simultaneously require its content to be \extype{all uppercase} and \extype{all lowercase}).
In the running example (Figure~\ref{fig:tree_example}, top panel), this stage commits keyword inclusion on ``report'' and ``policy'' plus a character cap at ``headline'', a list-length target at ``bullets'', and a word cap at each bullets[*].

\textbf{Parameter Backfill.} Each admitted constraint carries unresolved parameters, which the resolver binds in two stages.

\textit{Lexical resolution.} Lexical parameters draw their values from the per-task vocabulary pool produced by Keyword Mining. In the example (Figure~\ref{fig:tree_example}), the ``headline'' scope draws ``report'' and ``policy'' from the pool as its required keywords. Two consistency rules apply. First, since the required and forbidden words are drawn from the same pool, the sampler keeps them from colliding: no word is both required and forbidden, and neither is a substring of the other. Second, no keyword or identifier is drawn twice in the tree, so that every scope remains addressable by a unique alias.

\textit{Quantitative resolution.} Let $C, W, S, P$ denote the realized character, word, sentence, and paragraph counts of a scope, and write $X \in \{C, W, S, P\}$ with bounds $[X_{\min}, X_{\max}]$. Explicit constraints set one side at a time (\eg, \extype{at least $N$ words} sets $W_{\min} \gets N$), or both sides for an interval. Each lexical commitment lifts the corresponding lower bound. Let $T$ be the multiset of literal fragments forced by lexical commitments, where each required keyword contributes its literal and each pinned prefix, suffix, or postscript contributes its own. The resolver updates
\[
C_{\min} \gets \max\bigl(C_{\min},\, \textstyle\sum\nolimits_{t \in T} |t|\bigr),
\]
\[
W_{\min} \gets \max\bigl(W_{\min},\, \textstyle\sum\nolimits_{t \in T} \mathrm{wc}(t)\bigr),
\]
with $\mathrm{wc}(\cdot)$ denoting word count, and raises $S_{\min}$ and $P_{\min}$ by one for each pinned sentence- or paragraph-level fragment. Following readability studies~\cite{shannon1951prediction, kincaid1975derivation}, the resolver also enforces inter-unit ratio bands
\[
1 \le S/P \le 8,\;\; 3 \le W/S \le 40,\;\; 3 \le C/W \le 8,
\]
which propagate bounds across units (\eg, $C_{\max} \le 8\,W_{\max}$ and $C_{\min} \ge 3\,W_{\min}$). The resolver iterates the floor updates and band propagation to a fixed point, rejecting the draw if any $X_{\min} > X_{\max}$. A final coarse-to-fine pass (paragraphs first, characters last) pins each count to a value inside its surviving $[X_{\min}, X_{\max}]$. On the example, the character cap on ``headline'' sets $C_{\max}=60$, and the two required keywords joined by a space lift $C_{\min}$ from $0$ to $13$, leaving the feasible interval $[13, 60]$.

\textbf{Checker Generation.} The checker mirrors the finalized constraint tree node by node. Each constraint type has a corresponding hand-written deterministic checker. At each node it runs a boolean test on the slice of the response that belongs to that scope, and a container passes only when every child also passes on its own slice. Locating that slice is itself a parse step: each node first runs an extractor that descends from its parent's text to the region it governs, such as parsing the response as a JSON object, indexing into a list, or isolating the section under a heading, and the node's boolean test runs only after this extraction succeeds. If the response cannot be parsed down to a node's scope, the node fails and its entire subtree fails with it, so a constraint is never evaluated against the wrong span of text. For the running example (Figure~\ref{fig:tree_example}, bottom panel), the root parses the response as a JSON object and checks that its keys cover ``headline'' and ``bullets''. The ``headline'' value is then checked for the substrings ``report'' and ``policy'' and a $60$-character cap, the ``bullets'' value is checked for list length $3$, and each list item is checked for at most $15$ words. Each prompt in the released benchmark ships with its own deterministic checker.

\begin{figure}[t]
\centering
\refstepcounter{figure}\label{fig:tree_example}%
\begin{tcolorbox}[
  colback=green!8,
  colframe=green!55!black,
  title={\textbf{Figure~\thefigure: running example for~3.2}},
  fonttitle=\footnotesize,
  boxsep=1pt,left=4pt,right=4pt,top=2pt,bottom=2pt,
  fontupper=\footnotesize
]
\fvset{listparameters={\setlength{\topsep}{5pt}\setlength{\partopsep}{0pt}\setlength{\parsep}{0pt}\setlength{\itemsep}{0pt}}}
{\itshape Annotated constraint tree}
\begin{Verbatim}[fontsize=\scriptsize,frame=none,breaklines=false]
ROOT  (JSON object)
 +-- keys_required = {headline, bullets}
 +-- headline  (string, char_count <= 60)
 |    +-- must_include "report"
 |    +-- must_include "policy"
 +-- bullets  (list, list_length = 3)
      +-- bullets[*]  (string, word_count <= 15)
\end{Verbatim}

\smallskip
{\itshape Per-node checker (one-line Python predicate)}
\begin{Verbatim}[fontsize=\scriptsize,frame=none,breaklines=false]
root        : d = json.loads(x)
root.keys   : {"headline","bullets"} <= set(d)
headline    : "report" in s and "policy" in s
headline    : len(s) <= 60
bullets     : len(L) == 3
bullets[*]  : len(b.split()) <= 15
\end{Verbatim}
\end{tcolorbox}
\end{figure}

\subsection{Prompt Construction}
\label{sec:prompt-construction}

We convert the constraint tree into a single natural-language prompt by rendering each atomic constraint through a template and merging compatible sibling constraints into joint sentences.

\paragraph{Template Rendering.} Template Rendering converts each leaf constraint into one sentence. Consider the ``headline'' node in the running example (Figure~\ref{fig:tree_example}), which carries the constraint that the headline string must contain the keywords ``report'' and ``policy''. We pair every constraint type with a small set of templates, and the ``required keywords'' type maps to templates such as \emph{``\{alias\} must contain \{keywords\}.''}, where \{alias\} names the scope and \{keywords\} is the list of required words. Two slot values are then resolved. The alias is assigned by walking the tree top-down: the root is named ``the response'', and each child receives a phrase describing how it is reached from its parent (the value under a JSON key, an item at a position in a list, the section under a markdown heading). The JSON value under ``headline'' thus becomes ``the headline'', and the first element of the ``bullets'' list becomes ``item 1 of bullets''. The parameter slot is filled with the value committed during Parameter Backfill, here the list \{report, policy\}. Substituting both yields \ex{The headline must contain `report' and `policy'.}. The same procedure renders every other leaf in the example as a single sentence (\eg, \ex{The headline contains at most 60 characters.}).

\paragraph{Constraint Merging.} Constraint Merging collapses sibling sentences in the same scope into a single, more readable sentence. In the running example, the ``headline'' scope holds two rendered sentences after the previous step, \ex{The headline must contain `report' and `policy'.} and \ex{The headline contains at most 60 characters.}. We predefine a library of compact merge templates, where each template targets a specific combination of sibling constraint types. When the constraints in a scope match such a template, we apply it with probability $0.8$ and replace the separate sentences with the template's single sentence. Here the matched template rewrites the two sentences as \ex{The headline must include `report' and `policy' and contain at most 60 characters.}, the headline clause of the final prompt in Figure~\ref{fig:prompt_example}.

\begin{figure}[t]
\centering
\refstepcounter{figure}\label{fig:prompt_example}%
\begin{tcolorbox}[
  colback=violet!4,
  colframe=violet!75!black,
  title={\textbf{Figure~\thefigure: running example for~3.3}},
  fonttitle=\footnotesize,
  boxsep=1pt,left=4pt,right=4pt,top=2pt,bottom=2pt,
  fontupper=\footnotesize
]
Write a news brief based on the following article.
Return a \textcolor{violet!60!black}{JSON object} with the keys \textcolor{violet!60!black}{``headline''} and \textcolor{violet!60!black}{``bullets''}. The headline must include \textcolor{violet!60!black}{``report''} and \textcolor{violet!60!black}{``policy''} and contain at most \textcolor{violet!60!black}{60 characters}. The bullets field is a list of \textcolor{violet!60!black}{3 items}, each at most \textcolor{violet!60!black}{15 words}.

\smallskip
[article context omitted]
\end{tcolorbox}
\end{figure}
\section{Experiments}

\subsection{Experimental Setup}

\subsubsection{Models}
We evaluate seven models on \ourbenchmark, covering both leading proprietary APIs and open-weight checkpoints. Through hosted APIs we query \gptmodel~\cite{singh2025openai}, \claudemodel~\cite{anthropic2026claudeopus46}, \geminimodel~\cite{google2025gemini3}, \deepseekmodel~\cite{guo2025deepseek}, and \kimimodel~\cite{team2026kimi}. The remaining two open-weight models, \qwenmodel~\cite{yang2025qwen3} and \gemmamodel~\cite{deepmind2026gemma4}, run locally through the Ollama v0.21.0 runtime on a single NVIDIA RTX 4090 (24~GB) using its default 4-bit quantization.

We set the thinking effort of \gptanalysis and \claudeanalysis to high, and keep the provider default temperature because these endpoints do not permit temperature adjustment. All other models decode at temperature 0 for deterministic outputs.

\subsubsection{Evaluation Metrics}
Following IFEval~\cite{zhou2023instruction}, we report two strict accuracy metrics. Let $N$ denote the number of prompts in the benchmark, and let $C_i$ denote the set of constraints attached to prompt $i$. We write $\mathbf{1}[c]=1$ if the Python checker for constraint $c$ passes on the response and $0$ otherwise.

The \textbf{prompt-level accuracy} counts a response as correct only when every constraint is satisfied:
\begin{equation}
\mathrm{Acc}_{\mathrm{prompt}} = \frac{1}{N} \sum_{i=1}^{N} \prod_{c \in C_i} \mathbf{1}[c].
\end{equation}

The \textbf{instruction-level accuracy} averages over individual constraints rather than whole prompts:
\begin{equation}
\mathrm{Acc}_{\mathrm{inst}} = \frac{\sum_{i=1}^{N} \sum_{c \in C_i} \mathbf{1}[c]}{\sum_{i=1}^{N} |C_i|}.
\end{equation}

\subsection{Results}

\subsubsection{Overall Performance}

Table~\ref{tab:rq1_overall} reports prompt-level and instruction-level strict accuracy on \ourbenchmark, overall and stratified by constraint-tree depth.

\begin{table*}[t]
\centering
\small
{
\begin{tabular*}{\textwidth}{@{\extracolsep{\fill}}lccccccccccc@{}}
\toprule
\multirow{2}{*}{Model} & \multicolumn{5}{c}{Prompt-Level Accuracy (\%)} & & \multicolumn{5}{c}{Instruction-Level Accuracy (\%)} \\
\cmidrule(lr){2-6} \cmidrule(lr){8-12}
& Overall & $d{=}0$ & $d{=}1$ & $d{=}2$ & $d{=}3$ & & Overall & $d{=}0$ & $d{=}1$ & $d{=}2$ & $d{=}3$ \\
\midrule
\gptanalysis      & \textbf{53.7} & \textbf{86.7} & \textbf{49.3} & \textbf{43.3} & \textbf{35.3} & & \textbf{65.3} & \textbf{95.2} & 59.8          & \textbf{59.7}          & \textbf{66.9} \\
\claudeanalysis   & 42.8          & 78.7          & 42.0          & 28.0          & 22.7          & & 61.9          & 93.0          & \textbf{60.3} & 56.7 & 61.1          \\
\geminianalysis   & 33.2          & 74.7          & 38.0          & 14.0          & 6.0           & & 24.1          & 80.4          & 43.9          & 20.5          & 7.9           \\
\deepseekanalysis & 15.5          & 52.0          & 5.3           & 2.0           & 2.7           & & 37.7          & 71.0          & 42.4          & 35.7          & 31.4          \\
\kimianalysis      & 14.7 & 52.0 & 3.3  & 2.0  & 1.3  & & 33.0 & 78.7 & 28.2 & 22.7 & 35.1 \\
\qwenanalysis     & 14.8 & 48.7 & 7.3  & 2.0  & 1.3  & & 22.4 & 66.5 & 30.5 & 18.8 & 13.9 \\
\gemmaanalysis    & 19.8          & 63.3          & 10.0          & 2.0           & 4.0           & & 33.7          & 85.3          & 45.1          & 29.0          & 23.1          \\
\bottomrule
\end{tabular*}
}
\caption{Prompt-level and instruction-level accuracy on \ourbenchmark, overall and stratified by constraint-tree depth.}
\label{tab:rq1_overall}
\end{table*}

\textbf{\gptanalysis and \claudeanalysis lead at every depth.} \gptanalysis and \claudeanalysis, both run with thinking effort set to high, top the table on every cell: 53.7\% and 42.8\% prompt-level overall, far ahead of \geminianalysis at 33.2\% and a tightly bunched cluster of \gemmaanalysis (19.8\%), \deepseekanalysis (15.5\%), \qwenanalysis (14.8\%), and \kimianalysis (14.7\%). The ranking is preserved across $d{=}0$ through $d{=}3$, and the gap widens with depth. At $d{=}3$ \gptanalysis and \claudeanalysis retain 35.3\% and 22.7\% prompt-level accuracy, while every other model collapses to 6.0\% or below.

\textbf{Prompt-level accuracy decays steeply with depth.} Every model loses between 37 and 53 absolute points from $d{=}0$ to $d{=}1$, including the strongest two (\gptanalysis from 86.7\% to 49.3\%, \claudeanalysis from 78.7\% to 42.0\%). This universal cliff at the very first level of nesting suggests that the prompts current models see during instruction-tuning are dominated by flat constraint structures, and that hierarchical instruction-following remains a uniformly underdeveloped capability across today's leading models. Accuracy continues to fall as depth grows, but \gptanalysis and \claudeanalysis retain non-trivial pass rates at $d{=}3$ (35.3\% and 22.7\%), while the other five collapse to 6.0\% or below (\geminianalysis 6.0\%, \gemmaanalysis 4.0\%, \deepseekanalysis 2.7\%, \qwenanalysis 1.3\%, \kimianalysis 1.3\%). The persistence at depth for the two strongest models most likely traces to their advantage in parameter count and training-data volume, which together support generalization to instruction-following regimes that are thinly represented during training.

\textbf{The two metrics expose two distinct failure modes.} Four of the five lower-performing models exhibit scattered content slippage: \deepseekanalysis, \gemmaanalysis, \kimianalysis, and \qwenanalysis all show instruction-level accuracy well above prompt-level, with individual leaf constraints often passing but rarely all of them in the same response. Their leaf accuracy decays gradually with depth rather than collapsing, marking per-constraint slippage as the dominant error mode across these four. \geminianalysis shows the opposite signature: instruction-level (24.1\%) drops below prompt-level (33.2\%), and leaf accuracy falls off a cliff with depth. A failure at an outer container marks every descendant leaf as not applicable, so when \geminianalysis fails it tends to miss the structure entirely, dragging the entire subtree down. \gptanalysis and \claudeanalysis form a third pattern: the two metrics stay in a tight, positively correlated band, with instruction-level only 1.22$\times$ and 1.45$\times$ the prompt-level\ and leaf accuracy above 54\% at every depth.

\subsubsection{Failure Analysis}

We look at failure from two complementary angles: which constraint types account for the largest share of failures across all seven models, and how the two models that retain accuracy at $d{=}3$ plan their responses through the constraint hierarchy.

Table~\ref{tab:constraint_errors} ranks every constraint type by pooled pass rate across all seven models and reports each model's pass rate on the top five. The five hardest types are \emph{string\_length}, which fixes the response to an exact character count (53.5\%), \emph{num\_words}, which fixes the response to an exact word count (67.3\%), \emph{nth\_paragraph\_first\_word}, which fixes the first word of the $N$-th paragraph (73.2\%), \emph{value\_in\_set}, which restricts a value to a fixed allowed set (75.6\%), and \emph{num\_sentences}, which fixes the response to an exact sentence count (81.4\%).

Four of the five are numeric in nature, namely counting characters, counting words, counting sentences, and indexing the $N$-th paragraph, and they dominate the difficulty profile. This is also where the gap between the three best models and the other four is widest. On \emph{string\_length}, \gptanalysis, \claudeanalysis, and \geminianalysis pass 68\% to 90\% while the remaining four cluster at 24\% to 32\%. The pattern repeats on \emph{num\_words} (81\% to 97\% versus 37\% to 49\%) and on \emph{nth\_paragraph\_first\_word} (82\% to 98\% versus 31\% to 66\%). The hardest of the three is \emph{string\_length}, and one plausible explanation is exposure during instruction tuning: IFEval ships checkers for both word count and paragraph-indexed first word, but no checker for exact character length, so models likely receive far less training signal on character-precise length control.

\begin{table}[t]
\centering
\footnotesize
\setlength{\tabcolsep}{3pt}
\renewcommand{\arraystretch}{1.0}
\begin{tabular*}{\columnwidth}{@{\extracolsep{\fill}}lccccc@{}}
\toprule
             & string  & num    & nth para   & value  & num       \\
Model        & length  & words  & first word & in set & sentences \\
\midrule
Pooled pass (\%) & 53.5 & 67.3 & 73.2 & 75.6 & 81.4 \\
\midrule
\gptanalysis      & \textbf{90.1} & \textbf{96.7} & \textbf{97.8} & 82.4          & \textbf{98.3} \\
\claudeanalysis   & 68.2          & 87.2          & 96.4          & \textbf{88.2} & 95.7          \\
\geminianalysis   & 77.8          & 80.6          & 82.1          & 84.6          & 85.8          \\
\deepseekanalysis & 28.5          & 37.0          & 58.2          & 47.1          & 62.0          \\
\kimianalysis     & 29.8          & 48.6          & 40.4          & 77.8          & 68.6          \\
\qwenanalysis     & 24.1          & 47.8          & 30.7          & 71.4          & 66.9          \\
\gemmaanalysis    & 32.3          & 46.7          & 65.5          & 80.0          & 70.8          \\
\bottomrule
\end{tabular*}
\caption{Top five constraint types ranked by pooled failure rate across all seven models.}
\label{tab:constraint_errors}
\end{table}

Conversely, the two models that retain non-trivial accuracy at $d{=}3$, \gptanalysis and \claudeanalysis, both run with thinking effort set to high, survive precisely because they reason explicitly about the constraint tree before responding. We examine their reasoning traces to ask whether this planning step walks the tree branch-by-branch (depth-first) or level-by-level (breadth-first). On the 150 depth-3 prompts the two models trace in different styles. \gptanalysis writes topic-by-topic passages of the form ``I am thinking about how to satisfy X'', each focused on one specific constraint. \claudeanalysis instead opens almost every trace by restating the prompt's constraints as a numbered list (1., 2., 3., ...) before turning to how to satisfy each one. Because this enumeration is a verbatim echo of the prompt, we strip it from each \claudeanalysis trace. \gptanalysis traces need no such preprocessing.

For each constraint we pick a short distinctive phrase from its specification, such as the keys, keywords, or numbers it names, and record where that phrase first appears in the trace. Ordering constraints by these first-mention positions gives the sequence in which the model brought each one into its reasoning, and we label each prompt by whether the depth-first or the breadth-first traversal of the constraint tree matches that sequence. Both models lean depth-first. \claudeanalysis matches depth-first on 67\% of prompts and breadth-first on 33\%, while \gptanalysis matches depth-first on 62\% and breadth-first on 38\%. The depth-first lead widens for both models on prompts where the two canonical orderings are themselves well separated, ruling out an accidental near-tie. \claudeanalysis's DFS lead over BFS is about 1.4$\times$ that of \gptanalysis, in line with the qualitative trace styles.

\section{Conclusion}

Existing instruction-following benchmarks treat constraints as a flat checklist, leaving the hierarchical structure of real-world prompts unevaluated. We address this with \ourbenchmark, a benchmark of \dataamount prompts stratified across four constraint depths and built from \totalconstraintamount programmatically checkable constraint types. Evaluating seven leading models, we find that they handle flat ($d{=}0$) constraints well, but accuracy drops sharply as nesting deepens, with the strongest model only marginally exceeding 50\%. Hierarchical instruction following remains an open capability gap for today's strongest LLMs.

\section*{Limitations}

We highlight two limitations of \ourbenchmark. First, our prompts are synthesized from sampled constraint trees and template-driven renderings, so their phrasing is more uniform than naturally written prompts and may underrepresent the idiosyncratic styles users produce in practice. Second, several steps inside the task-setup preparation stage, including task pruning, vocabulary mining, and genre assignment, are themselves executed by large language models, so the quality of these intermediate artifacts inherits the underlying model's text-parsing and generation capabilities.

\bibliography{custom}

\appendix
\section{Appendix}

\subsection{LLM Contribution Statement}
We disclose that large language models were used as writing aids for: (i) improving grammar, fluency, and stylistic consistency (ii) rephrasing sentences for clarity and (iii) polishing LaTeX table code. No passages were adopted verbatim without manual review and editing by the authors.

\subsection{Prompt Template}

Two stages of the \ourbenchmark construction pipeline use LLM for analysis. We list both prompts in full below.

\paragraph{Format-type discovery.} Before we fix the structural constraint inventory of \ourbenchmark, we run a corpus-wide analysis to identify which output formats real instruction-following prompts already request. For every prompt in the source corpus, the model classifies its required output formats against a dynamically updated category list, extracts the source sentence that introduces each format, and proposes new categories when none of the known ones fit. The model additionally returns the nested structure of these format constraints as a tree, where a child constraint targets a sub-part introduced by its parent (for example, a JSON-with-key parent and a value-shape constraint on that key as its child), providing direct evidence that real prompts already specify formats hierarchically. The format families discovered here seed the root-shape choices used by the constraint sampler. The full prompt is shown in Figure~\ref{fig:prompt-format-discovery}.

\begin{figure*}[!t]
\begin{tcolorbox}[
  enhanced,
  boxrule=0.3pt, arc=1pt,
  left=3pt, right=3pt, top=3pt, bottom=3pt,
  boxsep=0pt,
  colback=gray!5, colframe=black!70,
  colbacktitle=black!70, coltitle=white,
  fonttitle=\bfseries,
  toptitle=1mm, bottomtitle=1mm,
  title={Format-type discovery prompt},
]
\begin{Verbatim}[fontsize=\fontsize{6pt}{7pt}\selectfont,formatcom=\normalfont\rmfamily,breaklines=true,breaksymbolleft={},breaksymbolright={}]
=== SYSTEM ===

You are an expert at analyzing output format requirements in LLM prompt templates.

**Definition of "format":** A format is the structural data format of the output -- how the output is organized and structured. Examples: JSON, Markdown, etc. Format is about the STRUCTURE of the output, NOT about the content. The following are NOT formats:
- Content constraints (e.g., "must include keyword X", "output must be a single word yes/no")
- String matching constraints (e.g., "use the exact phrase '...'")
- Tone/style constraints (e.g., "be concise", "use formal language")
- Numerical constraints (e.g., "under 100 words") unless they describe structural elements
- Fixed/canned responses (e.g., "output 'N/A' if not applicable")

Given a prompt template, your task is to:
1. Identify ALL output FORMAT types required by this prompt (structural formats only)
2. Classify each format into one of the known categories, OR propose a new category if none fits
3. Extract the EXACT sentence(s) from the prompt that define each format requirement
4. Analyze the hierarchical/nested structure of the format constraints as a tree:
   - Each constraint that operates on a sub-part defined by a parent constraint is a child node
   - There are NO predefined semantics for each level -- any constraint type can appear at any level
   - The nesting relationship is purely structural: a child constraint targets something defined/created by its parent constraint

Current known format categories (dynamically updated):
{categories_json}

Rules:
- "Format" means structural data format ONLY. Do not classify content-level constraints (string matching, fixed phrases, tone, word count, etc.) as formats.
- First check if the format fits an existing category. Only propose a new category if none fits.
- A prompt may have zero, one, or multiple format requirements.
- "No format requirement" is a valid answer if the prompt does not constrain output format. Many prompts have content constraints but no structural format constraint -- mark these as no_format_requirement = true.
- When proposing a new category, provide: name, description, and why existing categories don't fit. The new category must describe a structural data format, not a content constraint.
- For hierarchy analysis: represent constraints as a tree. A constraint is a child of another constraint if it applies to a sub-structure that the parent constraint defines. For example, if a parent says "output JSON with key 'items'", and a child says "'items' must be a list", the child is nested under the parent because it targets 'items' which the parent introduced. A child of that could be "each element in 'items' must have key 'name'" -- again targeting a sub-part of the parent. There is no limit on depth, and any constraint type (format, numerical, string, etc.) can appear at any level as long as it targets a sub-structure.

=== USER ===

Analyze the format requirements in the following complete prompt template. Focus on the FULL prompt text as the primary source of truth. The auxiliary annotations below are provided only as reference and may be incomplete.

---PROMPT START---
{prompt_text}
---PROMPT END---

[Auxiliary] Output Format/Style annotation: {output_format_style}
[Auxiliary] Extracted constraints (JSON): {extracted_constraints}

Return your analysis using the report_format_analysis function.
\end{Verbatim}
\end{tcolorbox}
\caption{Prompt for format-type discovery.}
\label{fig:prompt-format-discovery}
\end{figure*}

\paragraph{Task preparation.} For each base prompt drawn from the source corpus, we issue a single LLM call that (i) strips every pre-existing format, length, count, and style instruction, leaving only a bare content goal in the \texttt{task} field, (ii) classifies the task into one of twelve genres, (iii) proposes domain-specific vocabulary (\texttt{content\_hints}) that downstream samplers reuse as constraint parameters, (iv) detects and copies out any inline context the task operates on (such as a sentence to rewrite or a passage to summarise), (v) emits an audit list of every constraint string removed under step~(i) so the stripping pass is itself verifiable, and (vi) records the subset of root output shapes that would feel natural for the cleaned task. The cleaned task and content hints become the inputs to the constraint-tree sampler in the next stage. The full prompt is shown in Figure~\ref{fig:prompt-task-preparation}.

\begin{figure*}[!t]
\begin{tcolorbox}[
  enhanced,
  boxrule=0.3pt, arc=1pt,
  left=3pt, right=3pt, top=3pt, bottom=3pt,
  boxsep=0pt,
  colback=gray!5, colframe=black!70,
  colbacktitle=black!70, coltitle=white,
  fonttitle=\bfseries,
  toptitle=1mm, bottomtitle=1mm,
  title={Task preparation prompt},
]
\begin{Verbatim}[fontsize=\fontsize{6pt}{7pt}\selectfont,formatcom=\normalfont\rmfamily,breaklines=true,breaksymbolleft={},breaksymbolright={}]
=== SYSTEM ===

You annotate prompts for a constraint-following benchmark.

Your job for EACH input prompt:
1. Strip from the prompt EVERY token that pins the response's shape, length, count, format, medium, case, punctuation, or scaffold layout, leaving ONLY a bare content goal in `task`. Specifically remove:
   * EVERY explicit count or number that bounds output structure ("five Q&A pairs", "two paragraphs", "exactly 3 sections", "at least 100 words", "in 200 characters").
   * EVERY format / medium / structure word ("as JSON", "in HTML", "Markdown table", "bulleted list", "with headings", "in paragraphs", "wrapped in quotes", "all caps", "in lowercase", "with the title in bold").
   * EVERY scaffold / separator template / format example ("separated by ***", "using the format: Q: ... A: ...", "indicated by: * example bullet").
   * ALSO strip format names even when they appear as content nouns ("JSON schema", "HTML page", "Markdown document", "XML file", "CSV table", "YAML config", "SQL query"). Rephrase the task so the content goal stands alone WITHOUT naming any output format:
       "Write a JSON schema for a user profile with name, email, age."
         -> "Describe a user profile structure with name, email, age."
       "Generate an HTML page advertising a camera."
         -> "Write an advertisement for a camera."
       "Produce a Markdown tutorial on sorting."
         -> "Explain how to sort a list."
     The format intent is carried entirely by `allowed_root_formats` below; `task` must never name a format, in any role (directive OR content noun).
   The cleaned `task` reads like a plain content goal:
     "Provide question-and-answer pairs."
     "Write about banana peel biology."
     "List limericks about people named Bill."
   Keep genre / content-type nouns that are NOT format names (e.g., "advertisement", "poem", "recipe", "limerick", "email", "story") -- these describe what KIND of writing, not how it is rendered.
2. Classify the task into exactly one of 12 genres.
3. Propose domain-specific vocabulary (content_hints) that a constraint generator could plug into parameters like JSON key names, markdown heading titles, keyword requirements, etc. Hints must be semantically relevant to the task. ALL structural pools are MANDATORY (non-empty) regardless of genre -- the benchmark wraps every task in JSON / Markdown / Labeled root variants for hierarchical testing, so empty pools kill entire benchmark branches.
4. Detect inline `context`: material EMBEDDED IN THE PROMPT that the task operates ON, e.g., a sentence to rewrite, a passage to summarize, an email to reply to, a code snippet to refactor, a data row to describe, a list of items to reformat. If present, copy it VERBATIM into context.text (preserve quotes, line breaks, whitespace), set context.present=true, and pick the correct context.kind.

   `context` is ONLY text-to-be-processed by the model. The following are NOT context and must NOT be extracted as one:
   * A scaffold / format example / placeholder template ("* This is an example bullet", "Q&A #1 *** Q&A #2", "{ key: <value>, ... }" skeleton).
   * Format / structure instructions ("Put the names in bullet points", "Respond in JSON").
   * A repetition or rephrasing of the task itself.
   If the only embedded material is scaffold / instructions / task repetition, set context.present=false, text='', kind='none'. Self-contained generative prompts ('Write a poem about California', 'Suggest startup ideas') also have no context.
5. List each removed constraint verbatim for audit.
6. Decide `allowed_root_formats`: the subset of {string, json, list, markdown, labeled} that would feel natural as the response shape for THIS task. Use these definitions:
   * "string" -- free prose, no enforced structure. Natural for poems, lyrics, short open-ended answers, single-sentence rewrites, emails written as natural correspondence, stories.
   * "markdown" -- sectioned document with headings + paragraphs. Natural for essays, tutorials, articles, multi-section answers, anything that benefits from headings.
   * "list" -- the response is fundamentally an enumeration of items (rendered as a JSON array). Natural when the answer IS a list of options / suggestions / names / steps / ingredients.
   * "json" -- structured data with named fields. Natural for data records, product specs, schemas, comparisons with multiple attributes, anything where keyed access matters.
   * "labeled" -- sections written as "Label: content". Natural for recipes (Ingredients / Instructions), structured Q&A, parts of an analysis (Pros / Cons / Verdict).

   EXCLUDE any format that would feel absurd for this prompt's content. A poem must NOT allow "json" or "labeled". An email reply must NOT allow "list" or "json". A pure data-extraction task probably EXCLUDES "string". Include at least 2 formats when reasonable so the benchmark has variety, but never include a format that produces nonsense.

   SPECIAL CASE -- content intrinsically tied to one format: if the original prompt described content that IS a specific format (e.g., "Write a JSON schema...", "Generate an HTML page...", "Produce a Markdown tutorial..."), the format name has been stripped from `task` per rule 1, but the format intent must be preserved here. Set `allowed_root_formats` to that single format only (e.g., `["json"]` for the schema task, `["markdown"]` for the HTML / Markdown tutorial). Overriding the "at least 2 formats" guideline is correct in this case.

Rules:
- `task` must be a standalone, well-formed instruction -- the benchmark will append NEW formatting / count / structural constraints onto it, so it must not dictate any of those itself. Reject any task that still contains "exactly N", "as JSON", "in bullet points", "wrapped in quotes", "in paragraphs", etc.
- The `task` and the `context` are SEPARATE: `task` describes what to do (e.g., 'Rewrite the sentence in an unusual style.'); `context` holds the material to do it on. Do not concatenate them, do not paraphrase the context, and never put a scaffold / format example into context -- only real text-to-be-processed.
- Pool sizes (larger is better -- the benchmark samples without replacement). ALL pools below are MANDATORY for every genre.
    * suggested_keywords: 50-100 items (cover common terms + specific vocabulary; benchmark samples without replacement across ALL constraints)
    * suggested_json_keys: 10-25 items (snake_case; even prose tasks can be wrapped in JSON like {'title','intro','body','characters','theme',...})
    * suggested_headings: 10-20 items (any task can be sectioned; for a poem use ['Title','Opening Stanza','Imagery','Symbolism','Theme',...])
    * suggested_labels: 6-15 items (even casual answers can be grouped into ['Direct Answer','Reasoning','Caveats','Sources',...])
    * suggested_forbidden: exactly 2-3 items (keep TIGHT, high-frequency words)
- For `creative_writing` and `qa_answer`: do NOT skip structural pools. Imagine how a structured rewrite of the response would look and populate the keys / labels / headings accordingly. Empty pools are a bug.

=== USER ===

Prompt to annotate:

"""
{prompt}
"""

Return ONLY valid JSON matching the schema.
\end{Verbatim}
\end{tcolorbox}
\caption{Prompt for task preparation.}
\label{fig:prompt-task-preparation}
\end{figure*}

\subsection{Constraint Inventory}

Tables~\ref{tab:constraint-inventory} and~\ref{tab:constraint-inventory-cont} list all \totalconstraintamount constraints in \ourbenchmark: the \structureconstraintamount structure-level constraints that operate on container scopes (JSON, Markdown, list, and labeled-section regions) and the \valueconstraintamount content-level constraints that operate on string scopes. Each entry shows the constraint identifier, a short definition of what the deterministic Python checker verifies, and one of the natural-language templates from the prompt-construction stage.

\begin{table*}[!t]
\small
\setlength{\tabcolsep}{4pt}
\renewcommand{\arraystretch}{1.25}
\begin{tabular}{@{}p{0.95cm}p{3.6cm}p{4.7cm}p{6.1cm}@{}}
\toprule
Level & Constraint & Short definition & Example NL template \\
\midrule
 & json.has\_keys & The JSON object must contain at least the listed keys at its top level. & The response must include at least the following keys: \{keys\_joined\}. \\
 & json.key\_count & The number of top-level keys in the JSON object satisfies a relation against a target count. & The response must have \{rel\} \{n\} top-level keys. \\
 & array.length & The number of elements in the array satisfies a relation against a target count. & The response must have \{rel\} \{n\} elements. \\
 & array.elements\_at & Items at specified 0-indexed positions of the array are scoped as nested sub-shapes that carry further child constraints. & Items at positions \{positions\_joined\} of the response must satisfy the constraints below. \\
Struct. & list.items\_unique & The list contains no duplicate items. & All items in the response must be unique. \\
 & markdown.heading\_count & The Markdown document contains a target number of headings, optionally restricted to one heading level. & The response must include \{rel\} \{n\} headings\{level\_suffix\}. \\
 & markdown.has\_heading & The Markdown document contains a listed set of named headings (each opening a child sub-scope). & The response must include the following headings: \{headings\_joined\}. \\
 & markdown.has\_code\_block & The Markdown document contains at least one fenced code block of the given language, with no comments inside the code. & The response must contain at least one \{language\} code block. The code inside must not contain any comments. \\
 & labeled.has\_labels & The response contains a listed set of labeled sections, each label at line start followed by a colon. & The response must include at least the labels: \{labels\_joined\}, with each label placed at the start of a line followed by a colon. \\
 & labeled.section\_count & The response contains a target number of labeled sections under the same colon-suffixed convention. & The response must contain \{rel\} \{n\} labeled sections. Mark each section with a label followed by a colon at the start of a line. \\
\midrule
 & string.string\_length & The response length in characters falls within an inclusive [min, max] range. & The response must be between \{min\} and \{max\} characters long. \\
 & string.num\_words & The number of words in the response satisfies a relation against a target count. & The response must contain \{rel\} \{n\} words. \\
 & string.num\_sentences & The number of sentences in the response satisfies a relation against a target count. & The response must contain \{rel\} \{n\} sentences. \\
 & string.nth\_paragraph\_first\_word & The response has exactly $n$ paragraphs and the $i$-th paragraph begins with a fixed word. & The response must have exactly \{n\} paragraphs; the \{nth\_ord\} paragraph must begin with the word ``\{first\_word\}''. \\
 & string.contains & The response contains every item in a list of required substrings. & The response must contain the text: \{words\_joined\}. \\
 & string.not\_contains & The response contains none of a list of forbidden substrings. & The response must not contain any of: \{words\_joined\}. \\
Content & string.starts\_with & The response begins with a fixed prefix string. & The response must start with ``\{prefix\}''. \\
 & string.end\_checker & The response ends with an exact phrase, with nothing after it. & The response must end with the exact phrase ``\{end\_phrase\}''. No other words or characters should follow this phrase. \\
 & string.keyword\_frequency & Each listed keyword occurs in the response with a specified frequency. & The response must satisfy the following word counts: \{specs\_joined\}. \\
 & string.letter\_frequency & Each listed letter occurs in the response with a specified frequency. & The response must contain the letters as specified: \{specs\_joined\}. \\
 & string.capital\_word\_frequency & The number of fully capitalized words in the response satisfies a relation against a target count. & The response must contain \{rel\} \{n\} fully capitalized words. \\
 & string.capital\_letters & The response is written entirely in uppercase letters. & The response must be entirely in capital letters. \\
 & string.lowercase\_letters & The response is written entirely in lowercase letters. & The response must be entirely in lowercase. \\
\bottomrule
\end{tabular}
\caption{Constraint inventory of \ourbenchmark (Part 1 of 2). Lists the 10 structure-level constraints and the first 13 of 25 content-level constraints. The remaining 12 content-level constraints are listed in Table~\ref{tab:constraint-inventory-cont}.}
\label{tab:constraint-inventory}
\end{table*}

\begin{table*}[!t]
\small
\setlength{\tabcolsep}{4pt}
\renewcommand{\arraystretch}{1.25}
\begin{tabular}{@{}p{0.95cm}p{3.6cm}p{4.7cm}p{6.1cm}@{}}
\toprule
Level & Constraint & Short definition & Example NL template \\
\midrule
 & string.comma & The response contains no commas. & The response must not contain any commas. \\
 & string.quotation & The response is wrapped entirely in double quotes. & The response must be wrapped entirely in double quotes. \\
 & string.title & The response includes a designated title wrapped in double angle brackets. & The response must include the title <<\{title\_text\}>> wrapped in double angle brackets. \\
 & string.bullet\_list & The response contains exactly $n$ bullet points (using * or -). & The response must contain exactly \{n\} bullet points (using * or -). \\
 & string.section\_checker & The response is split into $n$ sections, each beginning with a fixed literal splitter. & The response must have \{n\} sections. Mark the beginning of each section with `\{splitter\}'. \\
Content & string.highlight\_section & The response contains at least $n$ inline highlighted passages wrapped in asterisks. & The response must contain at least \{n\} highlighted passages wrapped in asterisks, like *this*. \\
 & string.placeholder & The response contains at least $n$ bracketed placeholders of the form [NAME]. & The response must contain at least \{n\} placeholders in square brackets, like [NAME]. \\
 & string.postscript & The response ends with a postscript opened by a fixed marker and followed by exactly one sentence. & The response must end with a postscript starting with `\{marker\}', followed by exactly one sentence and nothing else. \\
 & string.value\_in\_set & The (sub-)value at this scope must equal one of an allowed value list. & The response must be one of: \{values\_joined\}. \\
 & string.constrained\_response & The whole response must be exactly one of an allowed value list. & Your answer must be exactly one of: \{values\_joined\}. \\
 & string.two\_responses & The response is two distinct outputs separated by the literal ******. & Produce two distinct responses separated by ******. \\
 & string.repeat\_prompt & The response first repeats the prompt verbatim and then answers it. & First repeat the prompt verbatim, then answer it. \\
\bottomrule
\end{tabular}
\caption{Constraint inventory of \ourbenchmark (Part 2 of 2). Lists the remaining 12 of 25 content-level constraints. The first 13 content-level constraints and all 10 structure-level constraints are listed in Table~\ref{tab:constraint-inventory}.}
\label{tab:constraint-inventory-cont}
\end{table*}

\subsection{Example Prompts by Constraint Depth}
\label{sec:depth-examples}
We show one representative \ourbenchmark prompt at each constraint depth, from a single flat constraint ($d{=}0$) to three levels of nested scopes ($d{=}3$).

\begin{tcolorbox}[enhanced,boxrule=0.3pt,arc=1pt,left=4pt,right=4pt,top=3pt,bottom=3pt,boxsep=0pt,colback=gray!5,colframe=black!70,colbacktitle=black!70,coltitle=white,fonttitle=\bfseries\footnotesize,toptitle=1mm,bottomtitle=1mm,title={Depth 0}]
\begin{Verbatim}[fontsize=\normalsize,formatcom=\normalfont\rmfamily,breaklines=true,breaksymbolleft={},breaksymbolright={}]
Write a poem about a curious cat.
Write the response in lowercase only.
\end{Verbatim}
\end{tcolorbox}

\begin{tcolorbox}[enhanced,boxrule=0.3pt,arc=1pt,left=4pt,right=4pt,top=3pt,bottom=3pt,boxsep=0pt,colback=gray!5,colframe=black!70,colbacktitle=black!70,coltitle=white,fonttitle=\bfseries\footnotesize,toptitle=1mm,bottomtitle=1mm,title={Depth 1}]
\begin{Verbatim}[fontsize=\normalsize,formatcom=\normalfont\rmfamily,breaklines=true,breaksymbolleft={},breaksymbolright={}]
Write a rap about an abyss.
Respond using Markdown formatting.
Include these top-level headings: "Imagery".
The section under heading 'Imagery' must be entirely in capital letters.
The response must include at most 3 headings.
\end{Verbatim}
\end{tcolorbox}

\begin{tcolorbox}[enhanced,boxrule=0.3pt,arc=1pt,left=4pt,right=4pt,top=3pt,bottom=3pt,boxsep=0pt,colback=gray!5,colframe=black!70,colbacktitle=black!70,coltitle=white,fonttitle=\bfseries\footnotesize,toptitle=1mm,bottomtitle=1mm,title={Depth 2}]
\begin{Verbatim}[fontsize=\normalsize,formatcom=\normalfont\rmfamily,breaklines=true,breaksymbolleft={},breaksymbolright={}]
Recommend a college with open enrollment and regional accreditation.
Respond with a JSON object.
The response must include at least the following keys: "transfer_credits".
The value of key 'transfer_credits' must include the following headings: "Pros and Cons".
The section under heading 'Pros and Cons' must be entirely in capital letters.
\end{Verbatim}
\end{tcolorbox}

\begin{tcolorbox}[enhanced,boxrule=0.3pt,arc=1pt,left=4pt,right=4pt,top=3pt,bottom=3pt,boxsep=0pt,colback=gray!5,colframe=black!70,colbacktitle=black!70,coltitle=white,fonttitle=\bfseries\footnotesize,toptitle=1mm,bottomtitle=1mm,title={Depth 3}]
\begin{Verbatim}[fontsize=\normalsize,formatcom=\normalfont\rmfamily,breaklines=true,breaksymbolleft={},breaksymbolright={}]
Write a riddle that describes the word 'key'.
Respond with a JSON object.
The response must include at least the following keys: "target_audience".
The value of key 'target_audience' must contain at least the keys "Sources".
The following positions in the value of key 'Sources' have specific requirements: 2.
Write item 2 of the value of key 'Sources' in lowercase only.
All items in the value of key 'Sources' must be unique.
Include exactly 3 entries in the value of key 'Sources'.
\end{Verbatim}
\end{tcolorbox}

\end{document}